\pdfoutput=1
\documentclass[letterpaper, 10 pt, conference]{ieeeconf}

\usepackage{times}

\usepackage{multicol}
\usepackage[bookmarks=true]{hyperref}
\usepackage{amsmath,amssymb,amsfonts}
\usepackage{algorithmic}
\usepackage{gensymb}
\usepackage{algorithm}
\usepackage{graphicx}
\usepackage{textcomp}
\usepackage{xcolor}
\usepackage{bm}
\usepackage{booktabs}
\IEEEoverridecommandlockouts

\newcommand{\norm}[1]{\left\| #1 \right\|}
\newcommand{\mat}[1]{\mathbf{#1}}
\newcommand{\vect}[1]{\bm{#1}}

\hypersetup{
   pdfauthor={Martin Peticco and Pulkit Agrawal},
   pdftitle={A Kinematic Metric for Fine Manipulation Ability in Robotic Hands},
   pdfsubject={Robotic Manipulation},
   pdfkeywords={Dexterous Manipulation;Robotic Hands;Kinematic Metric;Rolling Contact}
}

\begin{document}
\renewcommand{\arraystretch}{0.92}

\title{A Kinematic Metric for Fine Manipulation Ability in Robotic Hands}

\author{Martin Peticco and Pulkit Agrawal%
\thanks{The authors are with the Improbable AI Lab, Massachusetts Institute of Technology.
        {\tt\small \{mpeticco, pulkitag\}@mit.edu}}%
}

\maketitle

\begin{abstract}
Traditional robotic hand metrics focus on static properties such as workspace, manipulability, and grasp stability. However, these metrics do not directly measure dexterity under the standard definition in robotic manipulation: the ability to continuously change an object's pose within the hand while maintaining contact from an initial grasp. We introduce \emph{Kinematic Rolling Manipulation Ability} (KaRMA), a kinematic-only metric for fine manipulation that quantifies reachable in-hand translation and reorientation of a spherical test object within a two-finger precision pinch through feasible rolling motions. KaRMA enforces joint limits, collision constraints, rolling contact, and antipodal force feasibility, then explores reachable in-hand object poses via breadth-first search over translation and rotation primitives. KaRMA reports three scores: translational coverage (KaRMA-T), rotational coverage (KaRMA-R), and sensitivity to the initial grasp (KaRMA-S). We evaluate KaRMA on 16 widely used robotic hands and compare against static baselines, showing that KaRMA separates hands that rank identically under static proxies, reveals translation--rotation tradeoffs invisible to existing baselines, and is qualitatively consistent with selected published task benchmarks where Jacobian-based metrics can be misleading. Code and interactive demos are available at \href{https://martinpeticco.com/karma}{martinpeticco.com/karma}.
\end{abstract}

\IEEEpeerreviewmaketitle

\section{Introduction}
\label{sec:introduction}

A significant number of robotic hands and similar dexterous end-effectors have been presented in the past few years from both industry and academia, targeting fine manipulation and assembly tasks. Hand selection and design decisions are largely guided by simple proxies such as joint count and degrees of freedom (DOFs), which do not fully describe manipulation capability. For procurement and design iteration, it is useful to isolate the kinematic component of that capability in an objective score computed from the hand model alone.

Robotic manipulation literature often describes dexterity as the ability to continuously change an object's pose while maintaining contact from an initial grasp~\cite{BicchiHands2000,klein1987,bicchi2002}. Some metrics attempt to capture this, such as thumb opposition, fingertip workspace, or local manipulability~\cite{cotugno2014thumb,yoshikawa1985manipulability,interactivityfingers}. However, none directly quantifies a hand's ability to do this through a \emph{continuous} sequence of in-hand motions.

One could argue that a robot can simply reorient its end effector using its arm, or set down the object being manipulated to regrasp it in a new pose. However, this means giving up the last-mile precision that a dexterous end-effector affords~\cite{madollar2011}. Regrasping also increases task horizon, and many objects are not feasible to pick up immediately in a desired pose without fixtures~\cite{hollerbach1982workshop}. This matters in precision tasks such as assembly with small parts, fine reorientation during bin picking, or manipulating everyday objects in assistive settings. Thus, it is important to quantify how much fine object translation and reorientation a hand's kinematics permit, before any controller or sensing is layered on top.

This paper introduces \emph{Kinematic Rolling Manipulation Ability} (KaRMA), a kinematic metric for measuring fine manipulation capability of a robotic hand based on its kinematic model. KaRMA measures the reachable set of object translations and orientations for a small spherical test object in a precision pinch under rolling contact without regrasping or finger gaiting. The metric reports (i) a translational score, (ii) a rotational score, and (iii) a score that summarizes sensitivity to the initial pinch grasp. KaRMA is intended as a kinematics-only tool for hand procurement and kinematic design iteration: it compares kinematic structures, not complete hands with their fingertip geometry, friction, and control, and does not by itself predict task success. The contributions of this paper are:
\begin{enumerate}
    \item A kinematics-only dexterity metric based on reachability of in-hand object pose changes under rolling contact in a two-finger pinch.
    \item A practical computation procedure that evaluates translational ability, rotational ability, and seed robustness from a hand kinematic model under standardized constraints, including coupled joints.
    \item An evaluation of KaRMA across a set of widely used robot hands with comparisons to common baselines.
\end{enumerate}

\begin{figure*}[tp]
      \vspace{1.5mm}
      \centering
      \includegraphics[width=0.90\linewidth]{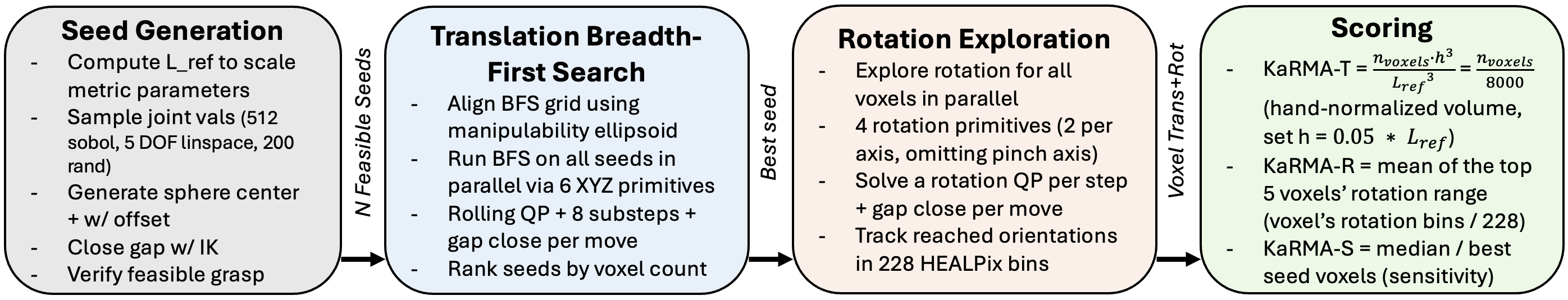}
      \vspace{-4mm}
      \caption{Overview of the KaRMA metric pipeline.}
      \label{fig:metric_overview}
      \vspace{-4mm}
\end{figure*}

\section{Related Work}
\label{sec:related_work}

\subsection{Robotic hand evaluation via kinematic proxies}
People often use simple quantitative descriptors such as DOF count and joint count to compare robot hands. Beyond this, kinematics-based proxies exist as well, such as the Kapandji score, which measures touching the thumb to different points on the hand~\cite{kapandji1986clinical}. Another such proxy is the measure of workspace volume for individual fingers~\cite{kerrroth1986volume}, which was built upon in Mouri et al.'s paper on the design of the Gifu Hand III, optimizing its design based on the intersection volume between the fingertip and thumb workspace volumes~\cite{Mouri2002AnthropomorphicRH}. You et al.'s ``Interactivity of Fingers'' extends this by measuring the intersection of multiple fingers~\cite{interactivityfingers}. Finally, local manipulability metrics such as the Yoshikawa index~\cite{yoshikawa1985manipulability} are commonly used in grasp quality analysis. These measures are useful for individual grasps and local configurations, but they do not directly quantify how much an object can be repositioned and reoriented within a grasp, per the commonly accepted definition of dexterity.

\subsection{Benchmarks and Task Suites}
Some benchmarks exist to directly capture object-centric in-hand capabilities through prescribed tasks and success criteria, such as the Elliott--Connolly benchmark~\cite{coulson2021elliott}, the Asterisk Test for two-finger translation~\cite{morrow2022asterisk}, and planar rotation benchmarks~\cite{morrow2023planarrotation}. More broadly, other general manipulation benchmarks involve performing a suite of tasks, often with learning-based methods, such as the Adroit dexterous manipulation suite~\cite{rajeswaran2018dexterous}. These benchmarks are useful for evaluating a full system, but their outcomes depend heavily on controllers, sensing, and training choices, making them less suitable for objectively comparing hand kinematics.

\subsection{Rolling contact formulation for robotic manipulation}
A natural way to isolate kinematics while still being object-based is to model in-hand manipulation through contact kinematics. Rolling-contact formulations exist that relate joint motion, contact motion, and object motion~\cite{montana1988kinematics}, which have been extended for use in dexterous motion generation and analysis. This has included classic rolling-based manipulation formulations and ``dexterous gripper'' designs that exploit rolling effects~\cite{bicchi1995rolling,bicchi2002}. More recent work prioritizes practical optimization or graph search methods for motion generation~\cite{sundaralingam2017ingrasp,cruciani2018dmg}. These formulations provide the modeling basis needed to define an objective, contact-constrained measure of in-hand object pose change from a kinematic hand model.

These works motivate a metric that goes beyond simple kinematic proxies, is less sensitive to system-level integration details, and measures realistic in-hand object pose manipulation via rolling contact constraints. KaRMA is proposed as a standardized kinematics-only score for thumb--index rolling-pinch manipulation, intended to complement rather than replace full task benchmarks.

\section{KaRMA Metric Definition}
\label{sec:metric_definition}

This section defines what KaRMA measures and what it reports. Details about computation are in Section~\ref{sec:computation}. A summary of the metric is seen in Fig.~\ref{fig:metric_overview}.

\subsection{Scope and Assumptions}
\label{subsec:scope}

KaRMA evaluates a two-finger precision pinch between the thumb and index finger on a small spherical test object under rolling contact and without regrasping or finger gaiting.

\textbf{Two-finger precision pinch.}
Thumb--index pinch is the most common grasp in fine manipulation and a lower-bound case for maintained-contact in-hand motion~\cite{feix2015grasp,bicchi2002,morrow2022asterisk}.

\textbf{Spherical test object.}
A sphere removes object-specific edge and corner effects, making the comparison object-agnostic and focused on rolling dexterity~\cite{borras_dimensional_2015}.

\textbf{Rolling contact, no regrasp.}
Maintained contact isolates kinematic in-hand capability and avoids dependence on controller design, gravity, or object geometry~\cite{BicchiHands2000,bicchi2002}.

\textbf{Kinematics only.}
KaRMA uses the hand kinematics only and is meant as a comparison tool, not to predict task success.

\subsection{Standardized Evaluation via $L_\mathrm{ref}$ Scaling}
\label{subsec:lref}

To make scores comparable across hand sizes and reproducible from a URDF, KaRMA non-dimensionalizes all length parameters (sphere radius, voxel size, tolerances) by scaling them with a characteristic length $L_\mathrm{ref}$. We compute $L_\mathrm{ref}$ from the URDF as
$L_\mathrm{ref} = \text{mxPr} + \text{mF}$,
where $\text{mxPr}$ is the maximum pairwise distance between finger knuckle joint origins (the first non-mimic revolute joint in each chain, advanced distally when that reduces $\text{mxPr}$), and $\text{mF}$ is the median knuckle-to-tip chain length (sum of link lengths along each chain). The sum captures both palm spread and finger reach, the two geometric factors that bound the envelope of feasible pinch locations. Parameters are defined at nominal $L_\mathrm{ref}=200\,\mathrm{mm}$ and then scaled by $L_\mathrm{ref}/200\,\mathrm{mm}$. Dimensionless parameters (friction coefficient, HEALPix resolution, BFS budget) are not scaled.

\subsection{Twist-Invariant Formulation}
\label{subsec:orientation}

In a two-finger pinch, rotation about the axis connecting the two contacts is uncontrollable (see Fig.~\ref{fig:metric_method}). We factor out this DOF by tracking the pinch-axis direction in the sphere's body frame. Since flipping the axis gives the same physical pinch, opposite directions are equivalent, leaving a two-dimensional orientation representation.

We discretize this space using HEALPix~\cite{gorski2005healpix}, which partitions $S^2$ into equal-area bins. With $n_\mathrm{side} = 6$ and antipodal identification, this yields $N_\mathrm{bin} = 6 n_\mathrm{side}^2 + 2 n_\mathrm{side} = 228$ bins at approximately $10^\circ$ resolution. Equal-area bins make the coverage fraction less sensitive to discretization bias across orientation space.

\subsection{Reported Scores}
\label{subsec:scores}
Given a kinematic model, KaRMA computes the reachable set of object translations and orientations from an initial feasible \emph{seed} pinch configuration under rolling contact. Three scores are reported:

\subsubsection{KaRMA-T (Translational Ability)}
The reachable set of sphere center positions is discretized on a voxel grid aligned with the principal directions of the seed configuration's manipulability ellipsoid. The translational score is the dimensionless ratio:
\begin{equation}
    \text{KaRMA-T} = \frac{n_\mathrm{voxels} \cdot h^3}{L_\mathrm{ref}^3}
    \label{eq:karma_t}
\end{equation}
where $n_\mathrm{voxels}$ is the number of reached voxels and $h$ is the (scaled) voxel edge length. Because the voxel scales with the hand ($h = 0.05\,L_\mathrm{ref}$), this reduces to $n_\mathrm{voxels}/8000$, the fraction of an $L_\mathrm{ref}$-sided cube filled by the reachable workspace; the count is already hand-normalized, so it is scale-invariant by construction rather than a raw voxel tally. We report KaRMA-T as the mean of this ratio over the three best-scoring seeds, which lowers the variance of the peak-capability estimate; the hand ranking is insensitive to this choice, with Spearman $\rho \geq 0.99$ across top-1, top-3, and top-5 aggregation. KaRMA-S separately reports how strongly the score depends on the seed.

\subsubsection{KaRMA-R (Rotational Ability)}
For each reached voxel, the fraction of the 228 orientation bins visited is recorded. KaRMA-R is the mean rotation coverage of the five best voxels:
\begin{equation}
    \text{KaRMA-R} = \frac{1}{|\mathcal{V}_5|}\sum_{v \,\in\, \mathcal{V}_5} \frac{|\mathcal{B}(v)|}{N_\mathrm{bin}}
    \label{eq:karma_r}
\end{equation}
where $\mathcal{V}_5$ is the set of the five reached voxels with highest orientation coverage and $\mathcal{B}(v)$ is the set of orientation bins reached at voxel $v$. We average over the top five rather than reporting a single-voxel maximum to reduce sensitivity to outlier voxels, while still focusing on the best region of the translational workspace. The score is averaged across the top 3 seeds, like KaRMA-T.

\subsubsection{KaRMA-S (Initial Grasp Sensitivity)}

KaRMA-S measures sensitivity to the initial grasp configuration. Each feasible seed is evaluated for its translational score, and KaRMA-S is the ratio of the median to the best:
\begin{equation}
    \text{KaRMA-S} = \frac{\text{KaRMA-T}_\mathrm{median}}{\text{KaRMA-T}_\mathrm{best}}
    \label{eq:karma_s}
\end{equation}
Intuitively, a KaRMA-S near 1 means most seeds perform similarly to the best seed, and near 0 means high performance is concentrated in specific configurations. KaRMA-S is defined by translational scores rather than rotational scores because translational reachability is more sensitive to the initial grasp, as different seeds produce different frontier voxels, directly affecting KaRMA-T. In contrast, KaRMA-R averages the top-five voxels' rotational coverage, and a seed that reaches even a few well-positioned voxels can still achieve high rotational coverage. Translational sensitivity is therefore a more conservative and informative robustness indicator. KaRMA-S should be interpreted jointly with KaRMA-T: a high KaRMA-S with low KaRMA-T means the hand uniformly lacks dexterity, as in very low-DOF hands.

\section{Full Metric Computation}
\label{sec:computation}

\begin{figure}[thpb]
      \vspace{1.5mm}
      \centering
      \includegraphics[width=0.78\linewidth]{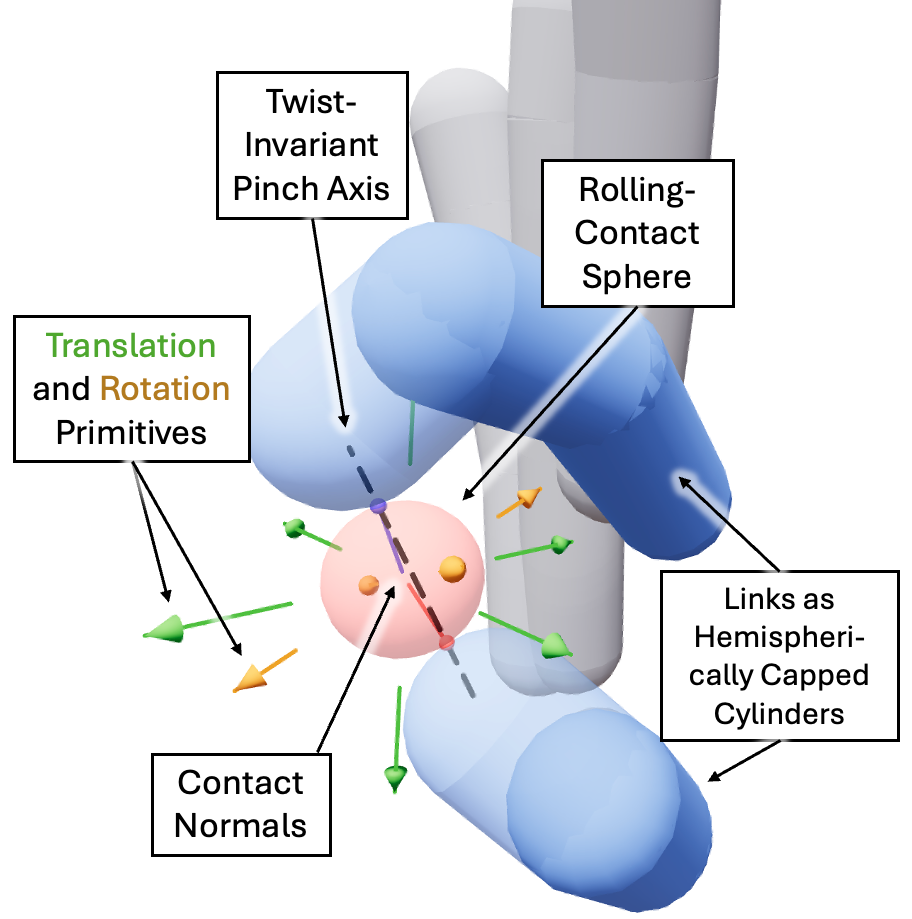}
      \caption{Schematic of the pinch grasp setup, including finger links, sphere, contact normals, BFS primitives, and pinch axis. ARMS model shown.}
      \label{fig:metric_method}
\end{figure}

This section describes how KaRMA is computed from a hand kinematic model. The pipeline has three stages: seed selection (\ref{subsec:seed}), two-phase BFS (\ref{subsec:bfs}), and scoring (\ref{subsec:scoring_computation}). All constraints are checked at every BFS step. See Fig.~\ref{fig:metric_method} for the setup.

\subsection{Contact Geometry}
\label{subsec:contact_geometry}

Each finger link is modeled as a capsule (cylinder with hemispherical caps) of fixed radius $r_\ell$, and the manipulated object is a sphere of radius $r_s$. Contact between a capsule and the sphere is checked using the distance from the sphere center $\vect{p}$ to the closest point $\vect{c}$ on the capsule axis segment:
\begin{equation}
    g = \norm{\vect{p} - \vect{c}} - r_s - r_\ell
    \label{eq:contact_gap}
\end{equation}
where $g$ is the signed contact gap. A contact is maintained when $|g| \leq \epsilon_g$, and the outward unit normal at contact is $\hat{\vect{n}} = (\vect{p} - \vect{c}) / \norm{\vect{p} - \vect{c}}$. At each BFS step, the contact pair is selected as the closest thumb link and closest index link to the sphere center, so the contacting link pair may change during BFS. For rolling constraints, we use the contact-point surface velocity Jacobian that relates joint motion and sphere rotation to contact-point velocity.

\subsection{Feasibility Constraints for Candidates}
\label{subsec:constraints}

\textbf{Contact gap:} both contacts $|g_i| \leq \epsilon_g$ after projection.

\textbf{Joint limits:} satisfy URDF bounds $q_j^- \leq q_j \leq q_j^+$.

\textbf{Collision avoidance:} the sphere does not penetrate any non-contact link capsule ($d - r_s - r_\ell \geq \epsilon_\mathrm{col}$), and thumb--index self-collisions satisfy a segment--segment distance of at least $2r_\ell + \epsilon_\mathrm{col}$ for all capsule pairs.

\textbf{Antipodal grasp feasibility:} contact normals satisfy a friction-cone force-closure test. For friction coefficient $\mu$, feasibility requires $\cos(\angle(-\hat{\vect{n}}_0, \hat{\vect{n}}_1)) \geq \cos(2 \arctan \mu)$.

\subsection{Rolling Contact QP}
\label{subsec:rolling_qp}

Each BFS primitive solves a quadratic program (QP) for a small local joint displacement $\Delta \vect{q}$ that approximately achieves either a target sphere translation $\Delta \vect{p}$ or a target rotation $\Delta \boldsymbol{\theta}_\mathrm{target}$ while maintaining locally rolling-consistent contact motion at both contacts.

For translation primitives, we optimize over $\vect{x} = [\Delta \vect{q}^\top,\, \Delta \boldsymbol{\theta}^\top]^\top$ and penalize deviation from ideal rolling (computed geometrically from $\Delta \vect{p}$) alongside regularization:
\begin{equation}
    \min_{\vect{x}} \norm{\mat{M} \vect{x} - \vect{b}}^2
    + \lambda_\theta \norm{\Delta \boldsymbol{\theta} - \Delta \boldsymbol{\theta}^*}^2
    + \lambda_r \norm{\Delta \vect{q}}^2
    \label{eq:rolling_qp}
\end{equation}
Following the rolling-contact kinematics of Montana~\cite{montana1988kinematics}, $\mat{M} = [\,\mat{J}_\mathrm{surf}\ \;{-}r\,[\hat{\vect{n}}]_\times\,]$ maps $(\Delta \vect{q}, \Delta \boldsymbol{\theta})$ to contact-point motion at both contacts, and $\vect{b} = \Delta \vect{p}$ is the commanded contact-point displacement. Rotation primitives are solved similarly with $\Delta \vect{p}=0$, targeting $\Delta \boldsymbol{\theta}_\mathrm{target}$ while maintaining rolling-consistent surface motion. Both QPs enforce joint step limits $|\Delta q_j| \leq \Delta q_\mathrm{max}$, joint position limits, and linearized gap constraints. Each voxel move is integrated using fixed sub-steps to maintain linearization accuracy, and a brief projection step re-closes residual contact gap. QPs are solved with OSQP~\cite{osqp2020}.

The objective in~\eqref{eq:rolling_qp} is a soft penalty, but a step is accepted only if its residual \emph{tangential} slip---sliding transverse to the rolling direction, excluding the free spin a sphere permits about the contact normal---stays below a hard gate placed at the elbow of the reach-versus-tolerance curve. This residual slip \emph{is} the local linearization error, and its median is below $0.5\%$ of the commanded motion for every hand above the low-DOF floor, so KaRMA measures essentially pure rolling wherever the score is informative. The five floor hands ($\dagger$) instead rely on substantial slip (median $11$--$32\%$) to move at all; a strict rolling equality would leave them with little reachable workspace, so the bounded gate keeps them on a common scale rather than reporting an uninformative zero.

\subsection{Seed Selection}
\label{subsec:seed}

The seed is the initial feasible thumb--index pinch configuration from which BFS starts. We deterministically generate candidate joint configurations using a fixed union of Sobol quasi-random samples, per-joint linspace sweeps, and uniform pseudo-random draws with a fixed seed. For each candidate, we propose a sphere center near the thumb--index contact midpoint (with small offsets) and project onto the two-contact manifold using a nonlinear least-squares IK solve with backtracking. Feasible seeds are then optionally polished by moving toward joint-range center in the nullspace of $\mat{J}_\mathrm{obj}=\frac{1}{2}(\mat{J}_{v,0}^\mathrm{surf}+\mat{J}_{v,1}^\mathrm{surf})$. All feasible seeds are evaluated with the full translational BFS.

The voxel grid is aligned with the seed configuration's manipulability ellipsoid: $\mat{R}_\mathrm{seed}$ is formed from the eigenvectors of $\mat{J}_\mathrm{obj}\mat{J}_\mathrm{obj}^\top$ (ordered by descending eigenvalue), with a deterministic sign convention and right-handedness enforced.

\subsection{Two-Phase BFS}
\label{subsec:bfs}

KaRMA uses a two-phase breadth-first search.

\textbf{Phase 1 (translation):} starting from the seed, BFS expands over voxels using six translation primitives ($\pm$ along each principal axis of the seed configuration). Each candidate step solves the rolling QP, applies projection, and checks all feasibility constraints. The search terminates when the frontier is exhausted or the state budget is reached.

\textbf{Phase 2 (rotation):} for each voxel from Phase~1, rotational primitives about the two controllable axes record reached HEALPix bins. KaRMA-R is the mean orientation coverage of the five best voxels.

\subsection{Scoring}
\label{subsec:scoring_computation}

After the BFS completes, KaRMA is computed as in Section~\ref{subsec:scores}.

\section{Experimental Setup}
\label{sec:experiments}

\subsection{Hands Evaluated}
We evaluate 16 hands: Ability, Allegro, D'Claw, Unitree Dex3, Unitree Dex5, Inspire, LEAP, OrcaHand, Sharpa, Shadow, Schunk SVH, Wuji, xHand1, xHandLite, plus anatomical human-hand skeletal models from ARMS and DLR~\cite{ARMSmodel2023,DLRhandmodel2013}. The set spans low to high DOF, compact to large designs, and includes non-anthropomorphic morphology.

\subsection{Default Parameters}
\label{subsec:default_params}
All evaluations use a sphere radius of $10\,\mathrm{mm}$ (at nominal $L_\mathrm{ref} = 200\,\mathrm{mm}$), voxel size $h = 10\,\mathrm{mm}$, friction coefficient $\mu = 0.6$, HEALPix $n_\mathrm{side} = 6$ ($228$ orientation bins at ${\approx}10^\circ$ resolution), and a BFS budget of $10{,}000$ states. Contact, collision, and step-size tolerances are fixed across hands and scaled with $L_\mathrm{ref} / 200\,\mathrm{mm}$ as described in Section~\ref{subsec:lref}.

\subsection{Computational Cost}
\label{subsec:compute_cost}
KaRMA is CPU-only and involves no learning. Evaluating all 16 hands on one 32-thread i9-13900KS takes under 18 minutes (6 seconds to 5 minutes per hand), under a $10{,}000$-state BFS budget with 42--670 candidate seed grasps per hand (median ${\approx}\,400$). A state is a reached workspace voxel, not a configuration-space sample, so cost does not grow exponentially with DOF, and the budget never binds---the largest hand reaches only 792 states.

\subsection{Baselines}
\label{subsec:baselines}

We compare KaRMA against kinematic proxies computed on the thumb and index chains. Coupled / mimic joints are taken into account everywhere.

\textbf{DOF count:} number of controllable joints.

\textbf{Total joint range:} sum of joint ranges (upper minus lower limit, in radians) over active thumb--index joints.

\textbf{Workspace volume:} convex hull volume of the fingertip workspace from 65{,}536 Sobol samples per finger, normalized by $L_\mathrm{ref}^3$. Fingers with $\leq 2$ DOF produce planar workspaces with zero 3D volume, making this baseline uninformative.

\textbf{Workspace intersection (opposability):} approximate volume of intersection of the thumb and index fingertip convex hulls, normalized by $L_\mathrm{ref}^3$~\cite{Mouri2002AnthropomorphicRH,interactivityfingers}.

\textbf{Yoshikawa manipulability:} mean manipulability of the combined thumb--index Jacobian over the same samples~\cite{yoshikawa1985manipulability}.

\textbf{Global conditioning index (GCI):} mean of $\sigma_\mathrm{min}/\sigma_\mathrm{max}$ over the samples, excluding near-singular configurations~\cite{gosselin1991gci}.

\subsection{Robustness and Sensitivity Experiments}
\label{subsec:robustness_experiments}

We validate KaRMA on five representative hands across the observed score range: LEAP (8 DOF, highest T), Shadow (9 DOF, highest DOF), xHand1 (6 DOF, mid-range), Dex3 (5 DOF, low T and DOF), and Inspire (3 DOF, lowest T of this set). All use a reduced $1{,}000$-state BFS budget.

\textbf{Invariance.} Test translation (four offsets), rotation (four rotations), and scale ($0.5\times$ and $2\times$ URDF scaling).

\textbf{Parameter sensitivity.} Independently perturb link lengths ($\pm 1\%$, $\pm 2\%$), joint limits ($\pm 1\degree$, $\pm 3\degree$), sphere radius ($\pm 2\%$, $\pm 5\%$), and capsule radius ($\pm 2\%$, $\pm 5\%$), and report the maximum absolute and relative change in KaRMA-T and KaRMA-R.

\textbf{Determinism.} Verify bit-for-bit reproducibility over three identical runs.

\begin{table*}[t]
\vspace{1.5mm}
\centering
\caption{KaRMA scores and selected kinematic baselines for all 16 hands, sorted by KaRMA-T. KaRMA-T and KaRMA-R are reported as means over the three best-scoring seeds. Workspace (WS) overlap is normalized by $L_\mathrm{ref}^3$. Hands with $\leq 2$ index DOF have zero 3D fingertip workspace intersection ($\dagger$).}
\label{tab:main_results}
\begin{tabular}{lcccccccccc}
\toprule
Hand & DOF & $L_\mathrm{ref}$ & Tot.\ Range & Thumb WS & Oppos. & Yoshikawa & GCI & \textbf{KaRMA-T} & \textbf{KaRMA-R} & \textbf{KaRMA-S} \\
     &     & (mm)             & (rad)       & $/L_\mathrm{ref}^3$ & $/L_\mathrm{ref}^3$ & (combined) & (combined) & & & \\
\midrule
LEAP       & 8 & 249.4 & 21.11 & 0.683 & 0.230 & 6.32e-8 & 0.102 & 0.097 & 0.304 & 0.45 \\
Allegro    & 8 & 222.5 & 12.59 & 0.221 & 0.053 & 9.87e-8 & 0.151 & 0.055 & 0.335 & 0.54 \\
D'Claw     & 6 & 327.2 & 16.25 & 0.315 & 0.117 & 2.60e-7 & 0.083 & 0.036 & 0.231 & 0.23 \\
Sharpa     & 9 & 188.3 & 12.74 & 0.283 & 0.070 & 4.34e-8 & 0.150 & 0.034 & 0.247 & 0.37 \\
Wuji       & 8 & 176.6 & 13.21 & 0.210 & 0.057 & 1.62e-8 & 0.117 & 0.034 & 0.233 & 0.35 \\
Shadow     & 9 & 184.8 & 12.64 & 0.178 & 0.033 & 1.00e-8 & 0.107 & 0.013 & 0.161 & 0.21 \\
ARMS       & 8 & 171.3 & 11.17 & 0.106 & 0.017 & 8.52e-9 & 0.115 & 0.012 & 0.149 & 0.34 \\
Dex5       & 8 & 185.9 & 11.91 & 0.354 & 0.036 & 9.86e-9 & 0.112 & 0.009 & 0.173 & 0.19 \\
DLR        & 9 & 204.5 & 14.15 & 0.176 & 0.012 & 1.18e-8 & 0.082 & 0.008 & 0.158 & 0.56 \\
Dex3$^\dagger$       & 5 & 155.3 &  8.80 & 0.308 & 0.000 & 1.09e-7 & 0.097 & 0.004 & 0.094 & 0.06 \\
xHand1     & 6 & 190.8 &  9.86 & 0.349 & 0.014 & 9.82e-9 & 0.091 & 0.003 & 0.097 & 0.09 \\
OrcaHand   & 7 & 168.9 & 12.69 & 0.259 & 0.007 & 5.58e-9 & 0.082 & 0.003 & 0.137 & 0.32 \\
xHandLite$^\dagger$  & 5 & 170.0 &  7.61 & 0.094 & 0.000 & 6.29e-8 & 0.091 & 0.002 & 0.061 & 0.06 \\
Inspire$^\dagger$    & 3 & 155.4 &  3.60 & 0.041 & 0.000 & 1.55e-3 & 0.208 & 0.0004 & 0.004 & -- \\
SVH$^\dagger$        & 4 & 182.9 &  4.09 & 0.036 & 0.000 & 1.88e-5 & 0.133 & 0.0004 & 0.010 & -- \\
Ability$^\dagger$    & 3 & 158.9 &  6.28 & 0.089 & 0.000 & 3.67e-4 & 0.376 & 0.0003 & 0.004 & -- \\
\bottomrule
\end{tabular}
\vspace{-4mm}
\end{table*}

\begin{table}[t]
\centering
\setlength{\tabcolsep}{4pt} 
\caption{KaRMA translational and rotational rankings for hands where the two scores diverge
($|\Delta| \geq 2$) or that rank in the top~5 for either score.
Overall T--R rank correlation is $\rho_s = 0.96$, confirming that both scores
capture related but non-identical kinematic properties.}
\label{tab:rank_divergence}
\begin{tabular}{lccccccc}
\toprule
Hand & DOF & T & T Rank & R & R Rank & $\Delta$ & Opp.\ Rk \\
\midrule
LEAP       & 8 & 0.097 & 1  & 0.304 &  2  & $-1$   & 1  \\
Allegro    & 8 & 0.055 & 2  & 0.335 &  1  & $+1$   & 5  \\
D'Claw     & 6 & 0.036 & 3  & 0.231 &  5  & $-2$   & 2  \\
Sharpa     & 9 & 0.034 & 4  & 0.247 &  3  & $+1$   & 3  \\
Wuji       & 8 & 0.034 & 5  & 0.233 &  4  & $+1$   & 4  \\
\midrule
ARMS       & 8 & 0.012 & 7  & 0.149 &  9  & $-2$   & 8  \\
Dex5       & 8 & 0.009 & 8  & 0.173 &  6  & $+2$   & 6  \\
Dex3       & 5 & 0.004 & 10 & 0.094 & 12  & $-2$   & 12 \\
OrcaHand   & 7 & 0.003 & 12 & 0.137 & 10  & $+2$   & 11 \\
\bottomrule
\end{tabular}
\end{table}

\section{Results}
\label{sec:results}

\subsection{KaRMA Scores}
\label{subsec:karma_results}

\begin{figure}[thpb]
      \centering
      \includegraphics[width=0.9\linewidth]{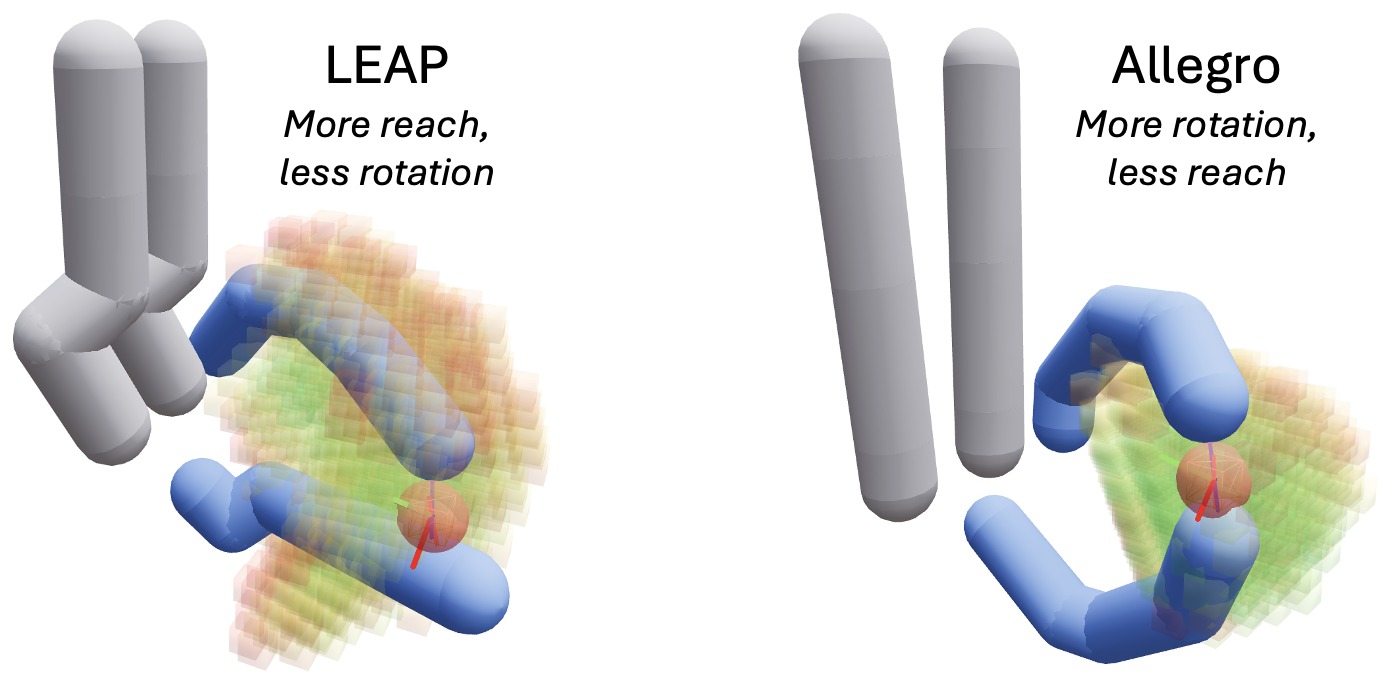}
      \caption{KaRMA on the two top-scoring hands (both 8 DOF). Voxel color denotes rotation coverage (red lowest, green highest). LEAP reaches $1.8\times$ the voxels (792 vs.\ 445), but over half of Allegro's voxels reach ${\geq}17\%$ orientation coverage versus 29\% for LEAP: the two hands swap ranks between KaRMA-T and KaRMA-R.}
      \label{fig:examples}
\end{figure}

Table~\ref{tab:main_results} reports KaRMA-T, KaRMA-R, and KaRMA-S for all 16 hands alongside kinematic baselines. KaRMA-T spans over two orders of magnitude, from 0.097 (LEAP) to 0.0003 (Ability), providing substantially more dynamic range than DOF count (3--9) or GCI (0.08--0.38) for discriminating between hands. The top five hands all achieve KaRMA-T $> 0.02$, whereas the bottom five are all 0.003 or below. Fig.~\ref{fig:examples} shows the two top-scoring hands: LEAP reaches the larger translational workspace, while Allegro sustains higher rotation coverage throughout its cloud. This is a distinction that other metrics do not capture. Reachable sets for all 16 hands, four case-study comparisons, and per-hand search diagnostics are collected in Appendices~\ref{app:gallery}--\ref{app:diagnostics}.

KaRMA-R ranges from 0.335 (Allegro) to 0.004 (Inspire, Ability), so the best hand reaches 33.5\% of the 228 HEALPix bins, averaged over five best voxels. Allegro ranks 2nd in T but 1st in R, while D'Claw ranks 3rd in T but only 5th in R. KaRMA distinguishes translational reach from reorientation coverage.

KaRMA-S measures initial-grasp sensitivity, a coarse robustness indicator. Allegro (0.54) and LEAP (0.45) are relatively stable across initial pinches, whereas xHand1 (0.09), Dex3, and xHandLite (both 0.06) concentrate their performance in a few grasps. DLR's high S (0.56) with low T is the caveat of Section~\ref{subsec:scores}. We omit KaRMA-S for the three hands below ten reachable voxels (Ability, Inspire, SVH), where it reduces to a ratio of single-digit integers.

KaRMA remains nonzero for the five zero-intersection hands ($\dagger$ in Table~\ref{tab:main_results}) because BFS measures reachability along the lower-dimensional manifold those fingers still traverse.

\subsection{Baseline Comparison}
\label{subsec:baseline_results}

Among baselines, workspace intersection correlates most strongly with KaRMA-T ($\rho_s = 0.93$, $p < 0.001$), which is expected since fingertip workspace intersection limits where feasible pinches can form. Total joint range ($\rho_s = 0.80$, $p < 0.001$) and DOF count ($\rho_s = 0.71$, $p = 0.002$) also correlate, but neither is reliable on its own: Shadow and DLR (both 9~DOF) rank only 6th and 9th in KaRMA-T, while D'Claw (6~DOF) ranks 3rd. Jacobian-based metrics show no correlation at all---Yoshikawa manipulability ($\rho_s = -0.20$, $p = 0.46$) and GCI ($\rho_s = -0.08$, $p = 0.76$)---because low-DOF hands like Ability have trivially well-conditioned Jacobians, giving them the highest GCI despite ranking last.

\begin{figure}[t]
\vspace{1.5mm}
\centering
\includegraphics[width=\columnwidth]{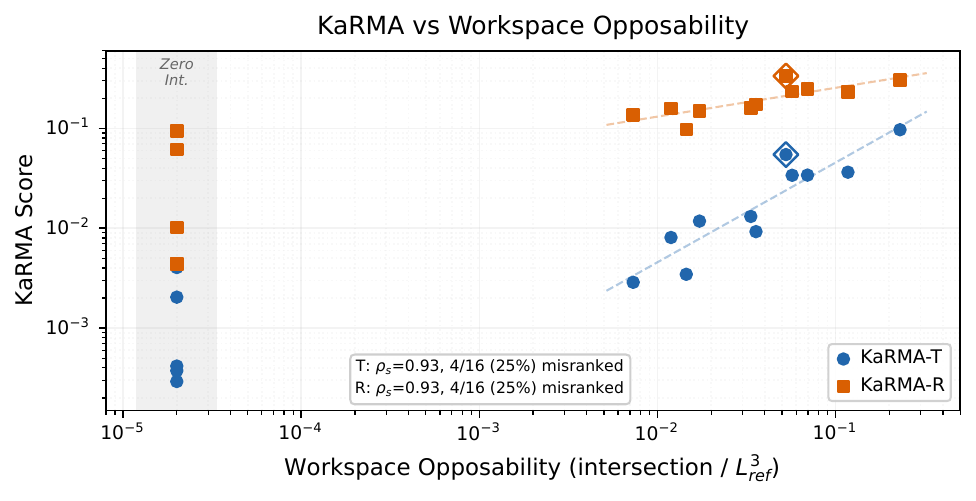}
\vspace{-3mm}
\caption{KaRMA-T (circles) and KaRMA-R (squares) vs.\ workspace opposability. The diamond marks Allegro, the one hand displaced $\geq$3 rank positions in both scores. Despite $\rho_s \geq 0.93$, it misranks 25\% of hands by $\geq$2 positions.}
\label{fig:karma_vs_opp}
\vspace{-4mm}
\end{figure}

The high $\rho_s$ with opposability is partly inflated by five zero-intersection hands clustering at the bottom of both rankings. Fig.~\ref{fig:karma_vs_opp} plots both KaRMA scores against opposability; despite the strong trend, rank displacement analysis reveals that opposability misranks 4 of 16 hands (25\%) by $\geq\!2$ positions in KaRMA-T, with 7\% of pairwise orderings inverted. The largest outlier (the diamond in Fig.~\ref{fig:karma_vs_opp}) is Allegro (opposability rank 5 $\to$ KaRMA-T rank 2), whose wide joint ranges sustain long rolling trajectories despite modest intersection. Dex5 (rank 6 $\to$ 8) and xHand1 (rank 9 $\to$ 11) are further $\geq\!2$-position examples: constraints reduce Dex5's moderate overlap to limited BFS reach, while joint limits eliminate 99\% of xHand1's unconstrained workspace (Table~\ref{tab:ablation}). Zero-intersection hands still score nonzero via rolling BFS, further separating the two measures.

Every baseline produces a single ranking; KaRMA produces three. Because KaRMA-S measures robustness, not capability, we focus on the T--R distinction. Table~\ref{tab:rank_divergence} shows that these two rankings, while correlated ($\rho_s = 0.96$), diverge in task-relevant ways. The rank difference $\Delta = \text{T rank} - \text{R rank}$ identifies hands relatively stronger at rotation ($\Delta > 0$) or translation ($\Delta < 0$). Among the top tier, Allegro ranks 1st in R but 2nd in T, while LEAP shows the opposite pattern. A researcher would pick Allegro to prioritize reorientation, or LEAP for repositioning. Opposability ranks both as ``high-overlap'' hands and cannot make this distinction. Mid-tier divergences follow the same pattern: Dex5 outperforms ARMS by three positions in R despite lower T, and OrcaHand ranks 12th in T but 10th in R. These distinctions are invisible to other metrics.

No baseline predicts the T--R divergence either: Spearman correlations between $\Delta$ and each baseline are all weak ($|\rho_s| \leq 0.31$, $p > 0.24$). Even with all baselines available, a practitioner cannot infer whether a hand is relatively better at translation or rotation without running KaRMA.

\textbf{Consistency with selected published benchmarks.}
Where prior task-based results exist, KaRMA and the published results are broadly aligned in the direction that simple kinematic proxies miss. Zhao et al.\ report that the Schunk SVH achieves better average dexterous manipulation performance than Inspire and Ability despite similar hand sizes~\cite{zhao2024dexmachina}. Jacobian-conditioning proxies can imply the opposite ordering for such low-DOF hands: GCI ranks Ability and Inspire above SVH (Table~\ref{tab:main_results}), and above most higher-DOF hands, because a low-dimensional Jacobian is trivially well-conditioned. KaRMA instead groups all three at the bottom of the table---KaRMA-R ranks SVH ($0.010$) above Inspire and Ability ($0.004$), as Zhao et al.\ report. Zhao et al.\ also report that XHand (xHand1 here) substantially outperforms Inspire on long-horizon articulated-object tasks~\cite{zhao2024dexmachina}; KaRMA similarly ranks xHand1 above Inspire for both translation and rotation, despite Jacobian proxies in Table~\ref{tab:main_results} suggesting the opposite. Finally, ISyHand reports stronger early-evaluation performance for Allegro than LEAP~\cite{isyhand2025}; KaRMA-R ranks Allegro 1st and LEAP 2nd, matching that early comparison even though total joint range would rank LEAP higher.

\subsection{Robustness}
\label{subsec:robustness_results}

Robustness experiments were conducted on the five representative hands.

\subsubsection{Invariance Properties}

Table~\ref{tab:invariance} summarizes the invariance results. KaRMA-T and KaRMA-R are exact under arbitrary world-frame translation and rotation and under $0.5\times$ and $2\times$ URDF scaling: every deviation in Table~\ref{tab:invariance} is zero, and three repeated runs per hand are bit-identical.

These invariances hold by construction. We canonicalize the base link to the world origin before forward kinematics, so a rigid translation or rotation of the robot cancels, and we solve the rolling QP and seed IK in non-dimensional form scaled by the sphere radius, so uniform scaling cancels. Because the metric counts voxels across discrete thresholds, we snap the canonicalized joint placements to a fixed $L_\mathrm{ref}$-relative grid, which removes the floating-point residual left by canonicalization that would otherwise flip boundary voxel and feasibility decisions.

\begin{table}[t]
\centering
\caption{Invariance of KaRMA-T and KaRMA-R. Each cell reports the maximum deviation as $T$ / $R$. Deviations for $T$ are in voxels; deviations for $R$ are in HEALPix bins. Baseline values are listed in the header row.}
\label{tab:invariance}
\setlength{\tabcolsep}{4pt}
\begin{tabular}{lccccc}
\toprule
 & LEAP & Shadow & xHand1 & Dex3 & INSP. \\
Baseline $T$ / $R$ & 792 / 69 & 110 / 37 & 32 / 22 & 34 / 22 & 4 / 1 \\
\midrule
Translation & 0 / 0 & 0 / 0 & 0 / 0 & 0 / 0 & 0 / 0 \\
Rotation    & 0 / 0 & 0 / 0 & 0 / 0 & 0 / 0 & 0 / 0 \\
Scale       & 0 / 0 & 0 / 0 & 0 / 0 & 0 / 0 & 0 / 0 \\
Determinism       & 0 / 0             & 0 / 0             & 0 / 0             & 0 / 0             & 0 / 0 \\
\bottomrule
\end{tabular}
\vspace{-3mm}
\end{table}

\subsubsection{Parameter Sensitivity}
We independently perturbed four model parameters: link lengths ($\pm 1$--$2\%$), joint limits ($\pm 1$--$3\degree$), sphere radius ($\pm 2$--$5\%$), and capsule radius ($\pm 2$--$5\%$), and recorded the maximum change in reached voxel count and rotation-bin coverage under the standard pipeline, which re-selects the best seed for each perturbed model. KaRMA is insensitive to modeling nuisances: for LEAP (792 baseline voxels), link geometry, sphere radius, and capsule radius each shift the voxel count by under 10\%, while its sensitivity is concentrated in joint limits (up to 20\% under $+3\degree$ widening), the constraint the ablation (Table~\ref{tab:ablation}) identifies as dominant. Relative changes are largest for small-workspace hands because a one-voxel change can represent a large percentage shift (e.g., Inspire's 4-voxel baseline means one voxel is 25\%). The largest relative deviation is xHand1 under $+3\degree$ joint-limit widening (34\%), followed by Shadow under $-3\degree$ narrowing (32\%). KaRMA-R follows similar patterns with generally smaller peak deviations (up to 33\% for xHand1); Dex3 is the exception, with comparable translational and rotational sensitivity (about 12\% each) rather than the translation-dominated pattern of the other hands.

To separate true search sensitivity from seed effects, we repeated all perturbations with the seed fixed to the baseline configuration. Under a fixed grasp, geometric perturbations (link lengths, radii) can produce larger, non-monotonic voxel changes, because a grasp tuned to the original geometry becomes mismatched after perturbation and flips boundary voxels. Re-selecting the seed in the standard pipeline attenuates this---often substantially---because the grasp is re-optimized for the perturbed model, which is why the reported sensitivities above stay small. Joint-limit perturbations, by contrast, reduce the fixed-seed reachable set directly, by $19$--$84\%$ under $3\degree$ narrowing, consistent with the constraint ablation (Table~\ref{tab:ablation}) identifying joint limits as the dominant constraint. With re-selection, no perturbation reorders the well-separated hands (only the near-tied xHand1/Dex3 pair can swap).

\subsection{Constraint Ablation}
\label{subsec:ablation}

To understand which constraints have the greatest effect, we performed a cumulative ablation study on the five representative hands, each on a single seed grasp held fixed across all levels, using the full 10{,}000-state budget with no constraint-level cap. Constraints were added in order: (i) contact gap only, (ii) $+$ joint limits, (iii) $+$ collision avoidance, (iv) full KaRMA ($+$ antipodal feasibility). Table~\ref{tab:ablation} shows both KaRMA-T and KaRMA-R at each level.

Joint limits are the dominant constraint for translation, reducing LEAP from 4{,}636 to 1{,}931 voxels ($-58\%$) and xHand1 from 6{,}250 to 58 ($-99\%$). Collision avoidance further reduces LEAP by $49\%$ but has minimal effect on hands already heavily constrained by joint limits. Antipodal feasibility removes an additional 14--40\% of the remaining configurations for the non-floor hands. For Inspire, all four voxels survive every ablation level, confirming its limitation is kinematic (QP infeasibility), not constraint-based.

KaRMA-R decreases or stays flat with each added constraint but with less dynamic range than KaRMA-T. For example, LEAP's KaRMA-T drops $5.9\times$ (0.580$\to$0.099) while KaRMA-R drops only $1.7\times$ (0.527$\to$0.304). This is expected: rotation coverage at each voxel depends on local joint mobility, which is less affected by the global workspace constraints that dominate translational reach. Notably, collision avoidance and antipodal feasibility have a proportionally larger effect on KaRMA-R than on KaRMA-T for several hands (e.g., xHand1's KaRMA-R drops $45\%$ from +Col to Full vs.\ $40\%$ for KaRMA-T), indicating that rotation exploration probes a wider range of configurations and is more likely to encounter grasp-infeasible states.

\begin{table}[t]
\vspace{1.5mm}
\centering
\caption{Cumulative constraint ablation results reported as $T$ / $R$. Conditions: (i) Gap only, (ii) +JL, (iii) +Col, (iv) Full KaRMA.}
\label{tab:ablation}
\setlength{\tabcolsep}{2pt}
\begin{tabular}{lcccc}
\toprule
 & (i) Gap & (ii) +JL & (iii) +Col & (iv) Full \\
\midrule
LEAP    & 0.580 / 0.527 & 0.241 / 0.401 & 0.123 / 0.332 & 0.099 / 0.304 \\
Shadow  & 0.509 / 0.527 & 0.016 / 0.214 & 0.016 / 0.173 & 0.014 / 0.161 \\
xHand1  & 0.781 / 0.473 & 0.007 / 0.193 & 0.007 / 0.177 & 0.004 / 0.097 \\
Dex3    & 0.007 / 0.343 & 0.004 / 0.097 & 0.004 / 0.097 & 0.004 / 0.094 \\
Inspire & 0.0005 / 0.004 & 0.0005 / 0.004 & 0.0005 / 0.004 & 0.0005 / 0.004 \\
\bottomrule
\end{tabular}
\vspace{-5mm}
\end{table}

\section{Discussion and Limitations}
\label{sec:discussion}

The results support two main points. First, KaRMA-T is broadly consistent with thumb--index overlap: hands with little fingertip overlap rarely translate the object far under maintained contact. At the same time, overlap alone is not sufficient. Workspace intersection ignores whether a continuous rolling trajectory exists under collisions, joint limits, and force-feasible antipodal contact. KaRMA filters overlap through these contact-consistency constraints, the source of many rank differences and most of the metric's practical value. The ablation (Table~\ref{tab:ablation}) confirms this: joint limits alone reduce unconstrained workspace by 58--99\% for the non-floor hands, and the full stack removes more, while KaRMA-R changes less across layers---rotation coverage is shaped more by local joint mobility than by the global constraints that dominate translation.

Second, KaRMA-R captures a manipulability axis that the baselines do not resolve. Allegro achieves the highest rotational coverage (0.335) despite ranking 2nd in translation, while D'Claw ranks 3rd in translation but only 5th in rotation. These divergences suggest hand selection should depend on the task: a design that repositions well need not be the best at reorientation, and no baseline predicts which way a hand will skew.

KaRMA should be interpreted as a standardized lower bound on thumb--index rolling-pinch dexterity, not as a universal dexterity score. It evaluates a single two-finger pinch on a sphere under maintained rolling contact, with no regrasp or finger gaiting and kinematics only, and depends on local linearization and discretization choices. These assumptions aid comparability but limit direct prediction of task success on arbitrary objects.

\section{Conclusion}
\label{sec:conclusion}

This paper introduced KaRMA, a kinematics-only metric for a standardized lower bound on fine manipulation ability in robotic hands. KaRMA measures the reachable set of object translations and orientations from a two-finger precision pinch on a spherical test object under rolling contact, reporting translational (KaRMA-T), rotational (KaRMA-R), and initial-grasp-sensitivity (KaRMA-S) scores. It is computed entirely from a URDF kinematic model, requires no controller or actuator specification, and is non-dimensionalized by a characteristic hand length so scores are comparable across hand sizes.

Across 16 robotic hands, KaRMA-T correlates strongly with workspace intersection ($\rho_s = 0.93$)---thumb--index overlap is the dominant factor in translation and a useful first-pass proxy---while KaRMA adds contact-consistency filtering that pure kinematic overlap cannot capture. KaRMA-R reveals an axis of manipulation ability that no baseline predicts: Allegro achieves the highest rotational coverage despite ranking only 2nd in translation, and although translational and rotational ability are strongly correlated overall, the mid-tier hands diverge in task-relevant ways. Jacobian-based metrics (Yoshikawa manipulability, GCI) show no correlation with either KaRMA score.

KaRMA currently evaluates only two-finger thumb--index pinch grasps; extending to multi-finger grasps and additional finger pairs would broaden its applicability. It could be useful in quantifying the advantages of having a robotic hand with certain biomimetic features, such as additional palm DOFs that give the pinkie and ring finger more dexterity. The spherical test object omits geometry-dependent effects, so adapting the metric to non-spherical objects is a natural extension. Finally, incorporating dynamic feasibility (torque limits, friction uncertainty) would move the metric closer to predicting task success, though at the cost of requiring more than a kinematic model.

\vspace{-1.5mm}

\section*{Acknowledgments}
We thank the Improbable AI Lab for helpful discussions and feedback. This research was supported in part by the Ministry of Trade, Industry, and Energy (MOTIE), Korea, under the ``Global Industrial Technology Cooperation Center program'' supervised by the Korea Institute for Advancement of Technology (KIAT) (Grant No.~P0028435). We were also supported by PEGATRON Corporation under the 2026--2030 MIT-PEGATRON Prometheus Research Collaboration Program, and by Analog Devices Inc.\ (ADI).

\appendices

\section{Reachable-Set Gallery}
\label{app:gallery}
\begin{figure*}[p]
  \centering
  \includegraphics[width=\textwidth]{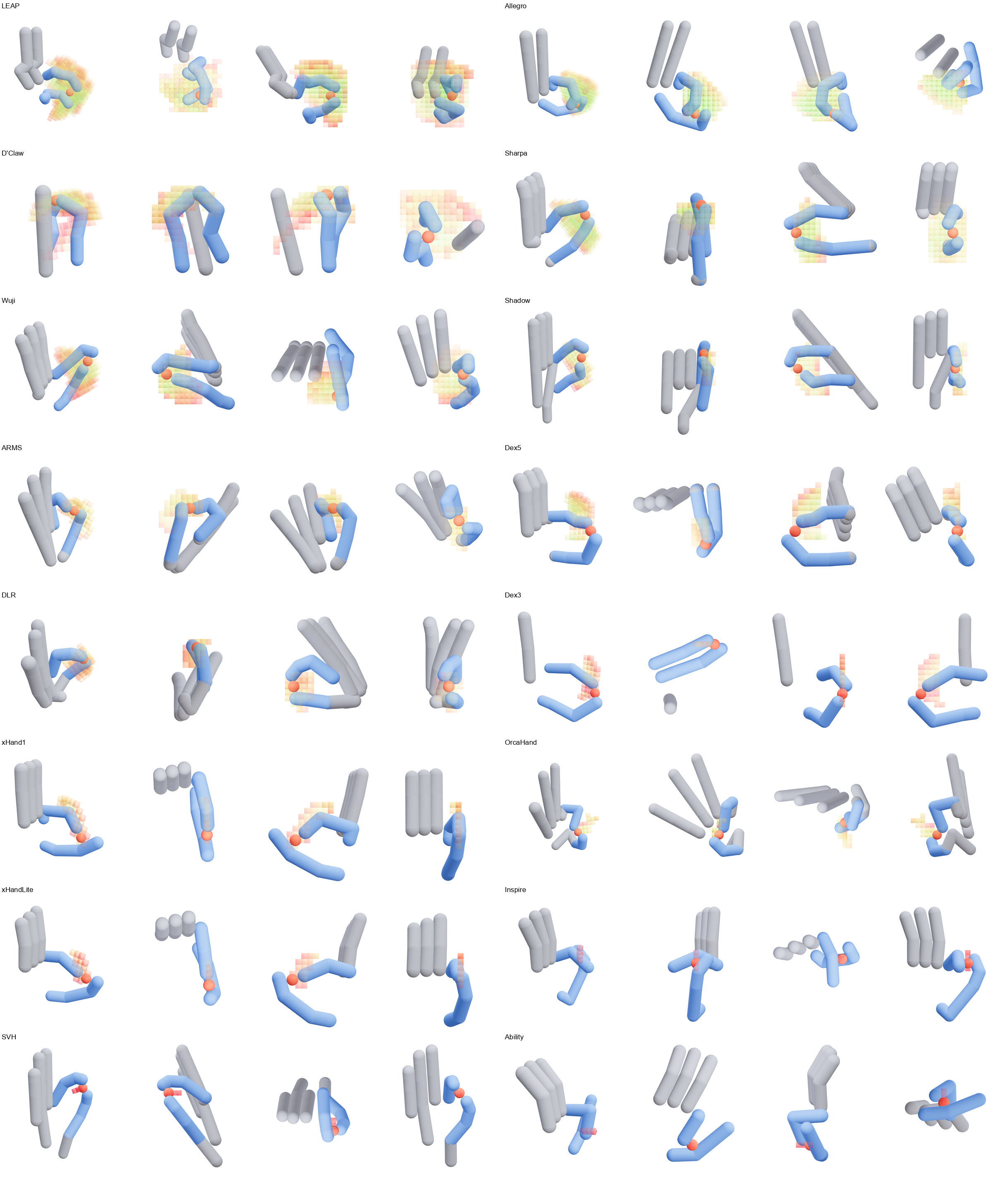}
  \caption{Reachable-set renders for all 16 hands, ordered by KaRMA-T (LEAP highest, Ability lowest). Each hand is shown from a perspective three-quarter \emph{hero} view and three orthographic views along the axes of the seed manipulability frame ($x$, $y$, $z$), the frame in which the translational voxel grid is defined. Voxels mark reached sphere-center positions, colored by rotation coverage of the $228$ orientation bins from red (low, $0\%$) through yellow to green (high, ${\geq}25\%$). Per-voxel opacity is scaled so sparse and dense clouds are comparably visible. Blue capsules are the thumb and index contact links, and gray capsules are the remaining links and fingers. The red sphere is the spherical test object. The orthographic axis views are given to better see the shape of the voxel clouds: we see high-scoring hands (LEAP, Allegro, D'Claw) fill a three-dimensional volume, whereas the 5-DOF hands (Dex3, xHandLite) collapse to a two-dimensional sheet, and the three lowest-scoring hands (Inspire, SVH, Ability) to just a few voxels in a line.}
  \label{fig:gallery}
\end{figure*}

Fig.~\ref{fig:gallery} shows the reachable sets KaRMA computes for all 16 hands, the same run used for Table~\ref{tab:main_results}.

\section{Case Studies}
\label{app:casestudies}
The following four comparisons illustrate example distinctions where KaRMA resolved some interesting aspect of the hands' abilities that static proxies missed. All statistics come from the same evaluation as Table~\ref{tab:main_results}. A voxel reorients well when its rotation coverage reaches at least $17\%$ of the 228 orientation bins. A set's shape is captured by the thickness of its thinnest axis in the seed-frame voxel grid and the largest fraction of its voxels in a single slice of that grid, which approaches 1 for a flat sheet.

\begin{figure}[t]
  \centering
  \includegraphics[width=\columnwidth]{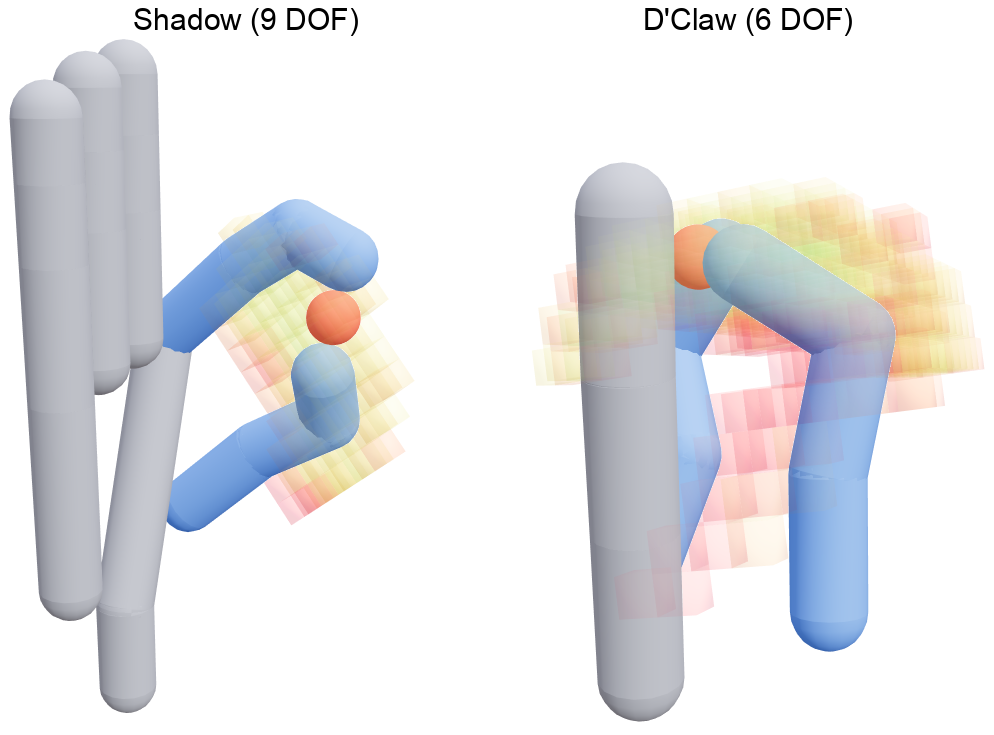}
  \caption{Shadow vs.\ D'Claw (hero views, shared scale). A 9-DOF anthropomorphic hand reaches a thin slab of 110 voxels, while a 6-DOF claw reaches a larger $315$-voxel volume.}
  \label{fig:cs_shadow_dclaw}
\end{figure}

\textbf{Shadow vs.\ D'Claw: DOF count alone does not predict reachable workspace.}
The 9-DOF Shadow Hand, long regarded as one of the most dexterous robotic hands, reaches only 110 voxels, while the non-anthropomorphic 6-DOF D'Claw reaches 315, nearly $3\times$ more (Fig.~\ref{fig:cs_shadow_dclaw}). Shadow's reachable set is a thin curved slab: its bounding box is $11{\times}3{\times}8$ voxels, only three voxels thick, with $49\%$ of the set in a single slice. D'Claw's is a full three-dimensional volume, $13{\times}11{\times}11$ with no slice holding more than $23\%$. D'Claw does, however, have a particularly irregularly shaped volume, notably there is a thin low-rotation strand where the sphere rolls down the side of one of the fingers. The constraint ablation (Table~\ref{tab:ablation}) identifies a cause for Shadow's weak performance: joint limits alone cut it from 4{,}070 to 128 voxels ($-97\%$), and the remaining constraints change little (128 to 110). Joint limits cost D'Claw only $78\%$ of its unconstrained workspace ($5{,}892$ to $1{,}280$ voxels), far less than Shadow's $97\%$, consistent with its wider joint range (16.3~rad against Shadow's 12.6). DOF count alone does not predict reachable workspace, consistent with its weaker correlation with KaRMA-T ($\rho_s = 0.71$) than that of total joint range ($\rho_s = 0.80$) in Section~\ref{subsec:baseline_results}.

\begin{figure}[t]
  \centering
  \includegraphics[width=\columnwidth]{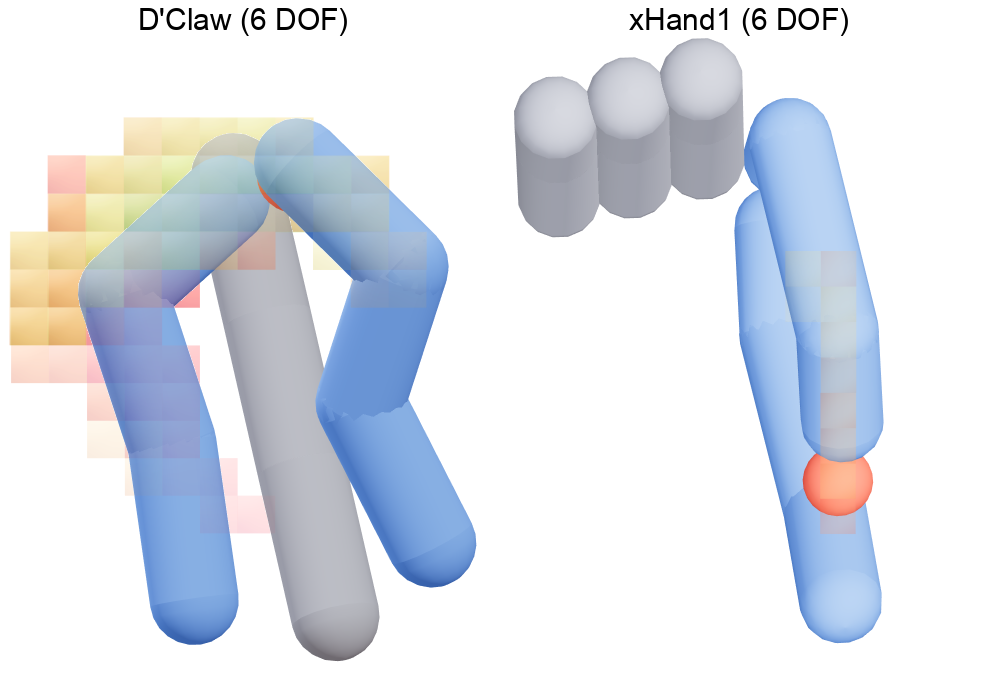}
  \caption{D'Claw vs.\ xHand1 (both seed $x$-axis, shared scale). At the same 6 DOF, D'Claw fills a volume while xHand1's sheet nearly vanishes edge-on.}
  \label{fig:cs_dclaw_xhand1}
\end{figure}

\textbf{D'Claw vs.\ xHand1: dimensional collapse at equal DOF.}
D'Claw and xHand1 are the only two 6-DOF hands in the study, yet their reachable sets have opposite dimensionality (Fig.~\ref{fig:cs_dclaw_xhand1}). D'Claw fills a volume, while xHand1's set is a near-planar sheet, $8{\times}2{\times}8$ voxels with $91\%$ of the set in a single slice, that nearly disappears when viewed edge-on along the seed $x$-axis. Both panels use that same view. The collapse is structural rather than a DOF effect: joint limits alone remove $99\%$ of xHand1's unconstrained workspace (Table~\ref{tab:ablation}, $6{,}250 \to 58$ voxels), and what survives the full constraint set is the 32-voxel sheet. By the same measure, joint limits cost D'Claw only $78\%$ against xHand1's $99\%$, consistent with its wider joint range (16.3~rad against xHand1's 9.9).

\begin{figure}[t]
  \centering
  \includegraphics[width=\columnwidth]{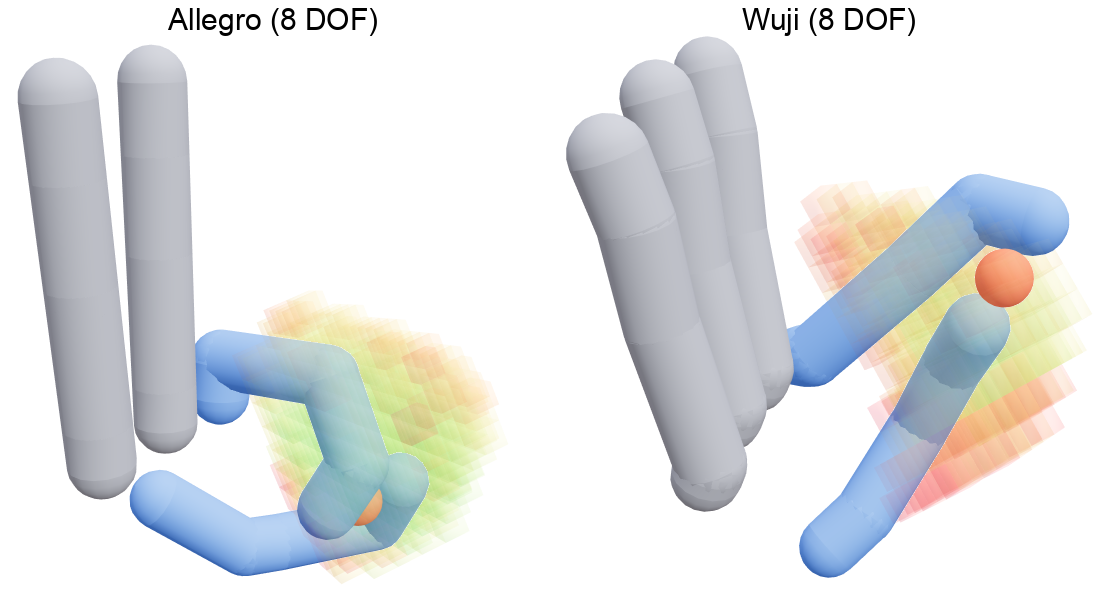}
  \caption{Allegro vs.\ Wuji (hero views, shared scale). Proxies rank them alike (opposability even favors Wuji), yet $53\%$ of Allegro's voxels reach at least $17\%$ rotation coverage against $9\%$ of Wuji's.}
  \label{fig:cs_allegro_wuji}
\end{figure}

\textbf{Allegro vs.\ Wuji: separating hands that proxies rank alike.}
Allegro and Wuji are nearly identical under the three proxies that correlate with KaRMA-T (Section~\ref{subsec:baseline_results}): both 8 DOF, total joint range 12.6 vs.\ 13.2~rad, and workspace opposability 0.053 vs.\ 0.057, which ranks Wuji slightly \emph{higher}. KaRMA separates them (Fig.~\ref{fig:cs_allegro_wuji}). At similar cloud size (445 vs.\ 310 voxels), $53\%$ of Allegro's voxels reach ${\geq}17\%$ orientation coverage versus $9\%$ for Wuji, and Allegro's mean per-voxel coverage is 0.174 vs.\ 0.101. Allegro ranks 1st in KaRMA-R and 2nd in T, while Wuji ranks 4th and 5th. None of the proxies anticipate this gap.

\begin{figure}[t]
  \centering
  \includegraphics[width=\columnwidth]{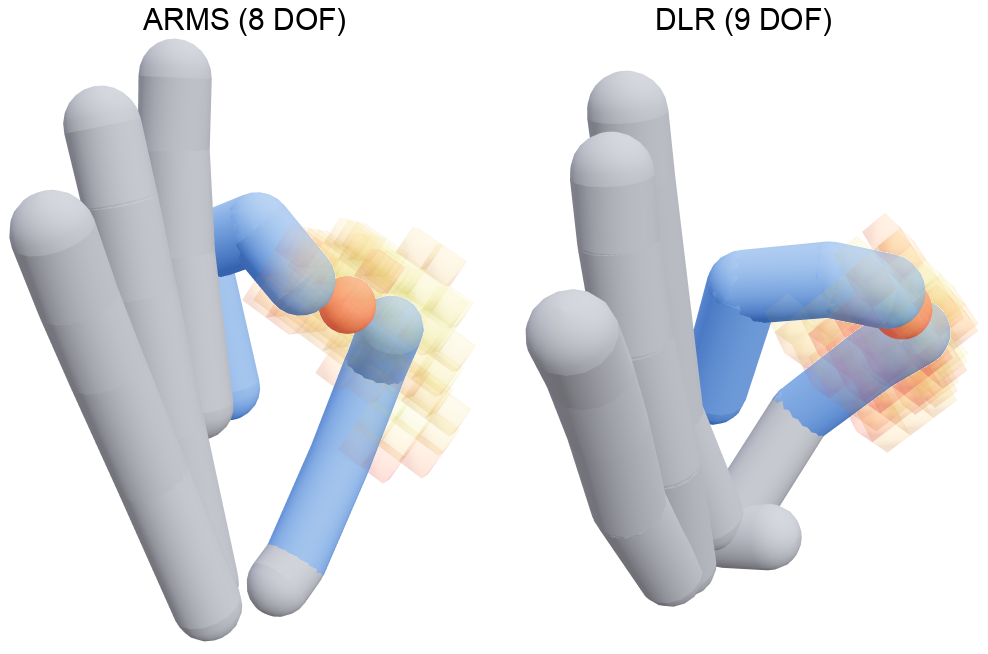}
  \caption{ARMS vs.\ DLR (hero views, shared scale). Two independent human-hand reconstructions score alike despite different DOF and size.}
  \label{fig:cs_arms_dlr}
\end{figure}

\textbf{ARMS vs.\ DLR: two human-hand models score alike.}
ARMS and DLR are anatomical skeletal models of the human hand, built independently by different research labs (built from measured joint motion vs.\ from MRI). The kinematic models themselves differ in various ways, such as in DOF (8 and 9) and size ($L_\mathrm{ref}$ 171 and 205~mm). Despite those differences, KaRMA scores and ranks them similarly (Fig.~\ref{fig:cs_arms_dlr}): KaRMA-T 0.012 vs.\ 0.008 and KaRMA-R 0.149 vs.\ 0.158. The fact that two independent models of the human hand score similarly suggests KaRMA reflects the hand's actual in-hand mobility, which the proxies do not capture. It also shows that anthropomorphism on its own does not yield the best in-hand workspace, as several robotic hands score higher than these human skeletal models.

\textbf{Where to grasp matters, and single-grasp reorientation is bounded.}
Two further findings follow from the per-voxel structure of the reachable set rather than from any single pair. First, for every hand with more than four reachable voxels, the voxel with the highest rotation coverage is not the seed voxel. The seed is selected for translational reach rather than rotation, so some gap is expected, but its size is not: Dex3's seed voxel reaches $0.9\%$ of the orientation bins while its best voxel reaches $11.4\%$, a $13\times$ difference, and Dex5's ratio is $7\times$. Where to hold an object for the best reorientation is hand-specific, and KaRMA's per-voxel map identifies that location, which no scalar metric can. Second, under the rotation exploration of Section~\ref{subsec:bfs}, no voxel of any hand reaches more than $35.5\%$ of the 228 pinch-axis bins (LEAP's best voxel, at 81), and KaRMA-R peaks at $33.5\%$ for Allegro. Even the most dexterous hand at its most capable position rotates the pinch axis into only about a third of the available directions through rolling from a fixed grasp. 

\section{Per-Hand Diagnostics}
\label{app:diagnostics}
Table~\ref{tab:diagnostics} reports per-hand search and contact diagnostics from the same evaluation as Table~\ref{tab:main_results}, supplementing the scores with quantities that characterize \emph{how} each hand reaches its set. Best and median voxel counts (over all feasible seeds) show how strongly capability concentrates in the best grasp, and the seed count is the number of feasible pinches found. Median $\alpha$ is the per-step tangential slip as a fraction of commanded motion (0 is pure rolling), and the gated-slip fraction is the share of accepted steps that touched the slip gate. The two slip columns separate the eleven medium-to-high-DOF hands (median $\alpha < 0.005$) from the five floor hands ($\dagger$, median $\alpha > 0.10$), which rely on having some amount of slip to move at all. The number of feasible pinches also does not indicate robustness across them, and across the 13 hands with a reported KaRMA-S the two run opposite ($\rho_s = -0.71$, $p = 0.007$): DLR yields only 42 feasible pinches yet has the highest KaRMA-S in Table~\ref{tab:main_results} (0.56), so its median grasp reaches over half of its best, though with a low KaRMA-T this is the uniformly modest case of Section~\ref{subsec:scores}. Dex3 yields 659 pinches but a KaRMA-S of 0.06, with its median grasp reaching 2 voxels against the best grasp's 34.

\begin{table}[t]
\centering
\caption{Per-hand search and contact diagnostics, ordered by KaRMA-T. Best / median voxels are over all feasible seeds, and seeds is the feasible-pinch count. Median $\alpha$ is median per-step tangential slip (fraction of commanded motion), and slip is the fraction of accepted steps at the slip gate. Floor hands ($\dagger$) rely on some amount of slip.}
\label{tab:diagnostics}
\setlength{\tabcolsep}{4pt}
\begin{tabular}{lccccccc}
\toprule
Hand & DOF & $L_\mathrm{ref}$ & Best & Med. & Seeds & Med. & Slip \\
     &     & (mm)             & vox  & vox  &       & $\alpha$ & frac \\
\midrule
LEAP & 8 & 249.4 & 792 & 355 & 254 & 0.0023 & 0.045 \\
Allegro & 8 & 222.5 & 445 & 241 & 513 & 0.0008 & 0.048 \\
D'Claw & 6 & 327.2 & 315 & 71 & 130 & 0.0029 & 0.082 \\
Sharpa & 9 & 188.3 & 290 & 107 & 408 & 0.0015 & 0.064 \\
Wuji & 8 & 176.6 & 310 & 109 & 228 & 0.0013 & 0.051 \\
Shadow & 9 & 184.8 & 110 & 23 & 594 & 0.0020 & 0.103 \\
ARMS & 8 & 171.3 & 96 & 33 & 401 & 0.0019 & 0.095 \\
Dex5 & 8 & 185.9 & 77 & 15 & 640 & 0.0022 & 0.103 \\
DLR & 9 & 204.5 & 72 & 40 & 42 & 0.0043 & 0.091 \\
Dex3$^\dagger$ & 5 & 155.3 & 34 & 2 & 659 & 0.1080 & 0.220 \\
xHand1 & 6 & 190.8 & 32 & 3 & 523 & 0.0033 & 0.137 \\
OrcaHand & 7 & 168.9 & 25 & 8 & 80 & 0.0029 & 0.155 \\
xHandLite$^\dagger$ & 5 & 170.0 & 17 & 1 & 670 & 0.1920 & 0.183 \\
Inspire$^\dagger$ & 3 & 155.4 & 4 & 1 & 258 & 0.2100 & 0.352 \\
SVH$^\dagger$ & 4 & 182.9 & 3 & 1 & 627 & 0.1218 & 0.408 \\
Ability$^\dagger$ & 3 & 158.9 & 3 & 1 & 291 & 0.3243 & 0.419 \\
\bottomrule
\end{tabular}
\end{table}

\bibliographystyle{IEEEtran}
\bibliography{references}

\end{document}